\documentclass[11pt,a4paper]{article}

\usepackage[]{algorithm2e}
\usepackage[hyperref]{acl2019}
\usepackage{times}
\usepackage{latexsym}
\usepackage[toc,page]{appendix}

\aclfinalcopy 

\usepackage[utf8]{inputenc} 
\usepackage[T1]{fontenc}    
\usepackage{hyperref}       
\usepackage{url}            
\usepackage{booktabs}       
\usepackage{amsfonts}       
\usepackage{amsmath}        
\usepackage{nicefrac}       
\usepackage{microtype}      
\usepackage{graphicx}
\usepackage{tipa}
\usepackage{listings}
\usepackage{booktabs}
\usepackage{paralist}
\usepackage{array}

\newcommand{\triple}[3] {\ensuremath{(\textit{#1},\textit{#2},\textit{#3})}}

\definecolor{nice-red}{rgb}{0.82, 0.1, 0.26}
\definecolor{nice-green}{rgb}{0.13, 0.55, 0.13}
\definecolor{nice-blue}{rgb}{0.13, 0.67, 0.8}

\newcommand{\emb}{\ensuremath{\mathbf{e}}}
\newcolumntype{x}[1]{>{\centering\arraybackslash\hspace{0pt}}p{#1}}

\usepackage[capitalise]{cleveref}

\newcommand{\ie}{\textit{i}.\textit{e}.}
\newcommand{\eg}{\textit{e}.\textit{g}.}

\title{NLProlog: Reasoning with Weak Unification for \\ Question Answering in Natural Language}

\author{
  Leon Weber\\
  Humboldt-Universit{\"a}t\\zu Berlin\\
  \texttt{weberple@hu-berlin.de} \\
  \And
  Pasquale Minervini\\
  University College London\\
  \texttt{p.minervini@cs.ucl.ac.uk} \\
  \And
  Jannes M{\"u}nchmeyer\\
  GFZ German Research Center\\for Geoscience Potsdam\\
  \texttt{munchmej@gfz-potsdam.de} \\
  \AND 
  Ulf Leser\\
  Humboldt-Universit{\"a}t\\zu Berlin\\
  \texttt{leser@informatik.hu-berlin.de} \\
  \And
  Tim Rockt{\"a}schel\\
  University College London\\
  \texttt{t.rocktaschel@cs.ucl.ac.uk} \\
}

\begin{document}

\maketitle

\begin{abstract}
Rule-based models are attractive for various tasks because they inherently lead to interpretable and explainable decisions and can easily incorporate prior knowledge. 
However, such systems are difficult to apply to problems involving natural language, due to its linguistic variability.
In contrast, neural models can cope very well with ambiguity by learning distributed representations of words and their composition from data, but lead to models that are difficult to interpret. In this paper, we describe a model combining neural networks with logic programming in a novel manner for solving multi-hop reasoning tasks over natural language.
Specifically, we propose to use a Prolog prover which we extend to utilize a similarity function over pretrained sentence encoders. We fine-tune the representations for the similarity function via backpropagation.
This leads to a system that can apply rule-based reasoning to natural language, and induce domain-specific rules from training data.
We evaluate the proposed system on two different question answering tasks, showing that it outperforms two baselines -- \textsc{BiDAF}~\citep{seo2016bidirectional} and \textsc{FastQA}~\citep{weissenborn2017fastqa} on a subset of the \textsc{WikiHop} corpus and achieves competitive results on the \textsc{MedHop} data set~\citep{welbl2017constructing}.
\end{abstract}

\section{Introduction}

We consider the problem of multi-hop reasoning on natural language data.
For instance, consider the statements ``\textit{Socrates was born in Athens}'' and ``\textit{Athens belongs to Greece}'', and the question ``\textit{Where was Socrates born?}''.
There are two possible answers following from the given statements, namely ``\textit{Athens}'' and ``\textit{Greece}''.
While the answer ``\textit{Athens}'' follows directly from ``\textit{Socrates was born in Athens}'', the answer ``\textit{Greece}'' requires the reader to combine both statements, using the knowledge that \emph{a person born in a city $X$, located in a country $Y$, is also born in $Y$.} 
This step of combining multiple pieces of information is referred to as \emph{multi-hop reasoning}~\citep{welbl2017constructing}.
In the literature, such multi-hop reading comprehension tasks are frequently solved via end-to-end differentiable (deep learning) models~\citep{sukhbaatar2015end, peng2015towards, seo2016query, raison2018weaver, henaff2016tracking, kumar2016ask, graves2016hybrid, dhingra2018neural}.
Such models are capable of dealing with the linguistic variability and ambiguity of natural language by learning word and sentence-level representations from data.
However, in such models, explaining the reasoning steps leading to an answer and interpreting the model parameters to extrapolate new knowledge is a very challenging task~\citep{46160,DBLP:journals/cacm/Lipton18,DBLP:journals/csur/GuidottiMRTGP19}.
Moreover, such models tend to require large amounts of training data to generalise correctly, and incorporating background knowledge is still an open problem~\citep{rocktaschel2015injecting,weissenborn2017dynamic,rocktaschel2017end,evans2017learning}.
In contrast, rule-based models are easily interpretable, naturally produce explanations for their decisions, and can generalise from smaller quantities of data.
However, these methods are not robust to noise and can hardly be applied to domains where data is ambiguous, such as vision and language~\citep{Moldovan2003-kh,rocktaschel2017end,evans2017learning}.
In this paper, we introduce \textsc{NLProlog}, a system combining a symbolic reasoner and a rule-learning method with distributed sentence and entity representations to perform rule-based multi-hop reasoning on natural language input.\footnote{\textsc{NLProlog} and our evaluation code is available at \url{https://github.com/leonweber/nlprolog}.}
\textsc{NLProlog} generates partially interpretable and explainable models, and allows for easy incorporation of prior knowledge.
It can be applied to natural language without the need of converting it to an intermediate logic form.
At the core of \textsc{NLProlog} is a 
backward-chaining theorem prover, analogous to the backward-chaining algorithm used by Prolog reasoners~\citep{Russell2010-sf}, where comparisons between symbols are replaced by differentiable similarity function between their distributed representations~\cite{sessa2002approximate}.
To this end, we use end-to-end differentiable sentence encoders, which are initialized with pretrained sentence embeddings~\citep{pagliardini2017unsupervised} and then fine-tuned on a downstream task. 
The differentiable fine-tuning objective enables us learning domain-specific logic rules -- such as transitivity of the relation \textit{is in} -- from natural language data.
We evaluate our approach on two challenging multi-hop Question Answering data sets, namely \textsc{MedHop} and \textsc{WikiHop}~\citep{welbl2017constructing}.
Our main contributions are the following:
\begin{inparaenum}[\itshape i)\upshape]
 \item We show how backward-chaining reasoning can be applied to natural language data by using a combination of pretrained sentence embeddings, a logic prover, and fine-tuning via backpropagation,
 \item We describe how a Prolog reasoner can be enhanced with a differentiable unification function based on distributed representations (embeddings),
 \item We evaluate the proposed system on two different Question Answering (QA) datasets, and demonstrate that it achieves competitive results in comparison with strong neural QA models while providing interpretable proofs using learned rules.
\end{inparaenum}

\section{Related Work} \label{sec:related}

Our work touches in general on weak-unification based fuzzy logic~\citep{sessa2002approximate} and focuses on multi-hop reasoning for QA, the combination of logic and distributed representations, and theorem proving for question answering.

\paragraph{Multi-hop Reasoning for QA.}
One prominent approach for enabling multi-hop reasoning in neural QA models is to iteratively update a query embedding by integrating information from embeddings of context sentences, usually using an attention mechanism and some form of recurrency~\citep{sukhbaatar2015end, peng2015towards,seo2016query,raison2018weaver}.
These models have achieved state-of-the-art results in a number of reasoning-focused QA tasks.
\citet{henaff2016tracking} employ a differentiable memory structure that is updated each time a new piece of information is processed.
The memory slots can be used to track the state of various entities, which can be considered as a form of temporal reasoning.
Similarly, the Neural Turing Machine~\citep{graves2016hybrid} and the Dynamic Memory Network~\citep{kumar2016ask}, which are built on differentiable memory structures, have been used to solve synthetic QA problems requiring multi-hop reasoning.
\citet{dhingra2018neural} modify an existing neural QA model to additionally incorporate coreference information provided by a coreference resolution model.
\citet{de2018question} build a graph connecting entities and apply Graph Convolutional Networks~\citep{kipf2016semi} to perform multi-hop reasoning, which leads to strong results on \textsc{WikiHop}.
\citet{zhong2019coarse} propose a new neural QA architecture that combines a combination of coarse-grained and fine-grained reasoning to achieve very strong results on \textsc{WikiHop}.

All of the methods above perform reasoning implicitly as a sequence of opaque differentiable operations, making the interpretation of the intermediate reasoning steps very challenging.
Furthermore, it is not obvious how to leverage user-defined inference rules during the reasoning procedure.
\paragraph{Combining Rule-based and Neural Models.}
In Artificial Intelligence literature, integrating symbolic and sub-symbolic representations is a long-standing problem~\citep{besold2017neural}.
Our work is very related to the integration of Markov Logic Networks~\citep{DBLP:journals/ml/RichardsonD06} and Probabilistic Soft Logic~\citep{DBLP:journals/jmlr/BachBHG17} with word embeddings, which was applied to Recognizing Textual Entailment (RTE) and Semantic Textual Similarity (STS) tasks~\citep{garrette2011integrating, garrette2014formal, beltagy2013montague, beltagy2014probabilistic}, improving over purely rule-based and neural baselines.
An area in which neural multi-hop reasoning models have been investigated is Knowledge Base Completion (KBC)~\citep{das2016chains,cohen2016tensorlog,neelakantan2015compositional,rocktaschel2017end,das2017go,DBLP:journals/jair/EvansG18}.
While QA could be in principle modeled as a KBC task, the construction of a Knowledge Base (KB) from text is a brittle and error prone process, due to the inherent ambiguity of natural language.
Very related to our approach are Neural Theorem Provers (NTPs)~\citep{rocktaschel2017end}: given a goal, its truth score is computed via a continuous relaxation of the backward-chaining reasoning algorithm, using a differentiable unification operator.
Since the number of candidate proofs grows exponentially with the length of proofs, NTPs cannot scale even to moderately sized knowledge bases, and are thus not applicable to natural language problems in its current form.
We solve this issue by using an external prover and pretrained sentence representations to efficiently discard all proof trees producing proof scores lower than a given threshold, significantly reducing the number of candidate proofs.
\paragraph{Theorem Proving for Question Answering.}
Our work is not the first to apply theorem proving to QA problems.
\citet{Angeli2016-jw} employ a system based on Natural Logic to search a large KB for a single statement that entails the candidate answer.
This is different from our approach, as we aim to learn a set of rules that combine multiple statements to answer a question.
Systems like Watson~\citep{Ferrucci2010-ta} and COGEX~\citep{Moldovan2003-kh} utilize an integrated theorem prover, but require a transformation of the natural language sentences to logical atoms.
In the case of COGEX, this improves the accuracy of the underlying system by 30\%, and increases its interpretability.
While this work is similar in spirit, we greatly simplify the preprocessing step by replacing the transformation of natural language to logic with the simpler approach of transforming text to triples by using co-occurences of named entities.
\citet{Fader2014-nq} propose \textsc{OpenQA}, a system that utilizes a mixture of handwritten and automatically obtained operators that are able to parse, paraphrase and rewrite queries, which allows them to perform large-scale QA on KBs that include Open IE triples.
While this work shares the same goal -- answering questions using facts represented by natural language triples -- we choose to address the problem of linguistic variability by integrating neural components, and focus on the combination of multiple facts by learning logical rules.%

\section{Background} \label{sec:weak_unification}
In the following, we briefly introduce the backward chaining algorithm and unification procedure~\citep{russell2016artificial} used by Prolog reasoners, which lies at the core of \textsc{NLProlog}.
We consider Prolog programs that consists of a set of rules in the form of Horn clauses:
\begin{equation*}
\begin{aligned}
    h(& f^h_1, \ldots, f^h_n)\ \Leftarrow\ \\
    & p_1(f^1_{1}, \ldots, f^1_m)\ \wedge\ \ldots\ \wedge\ p_B(f^B_{1}, \ldots, f^B_{l}),
\end{aligned}
\end{equation*}
where $h, p_i$ are predicate symbols, and $f^i_j$ are either function (denoted in lower case) or variable (upper case) symbols.
The domain of function symbols is denoted by $\mathcal{F}$, and the domain of predicate symbols by $\mathcal{P}$.
$h(f^h_1, \ldots, f^h_n)$ is called the \emph{head} and $p_1(f^1_{1}, \ldots, f^1_m)\ \wedge\ \ldots\ \wedge\ p_B(f^B_{1}, \ldots, f^B_{l})$ the \emph{body} of the rule.
We call $B$ the \emph{body size} of the rule and rules with a body size of zero are named \emph{atoms} (short for \emph{atomic formula}). 
If an atom does not contain any variable symbols it is termed \emph{fact}.
For simplicity, we only consider function-free Prolog in our experiments, \ie{} Datalog \citep{gallaire1978logic} programs where all function symbols have arity zero and are called \emph{entities} and, similarly to related work~\citep{sessa2002approximate, Julian-Iranzo2009-xa}, we disregard negation and disjunction.
However, in principle \textsc{NLProlog} also supports functions with higher arity.
A central component in a Prolog reasoner is the \emph{unification} operator: given two atoms, it tries to find variable substitutions that make both atoms syntactically equal.
For example, the atoms $\textit{country}(\textit{Greece}, \textit{Socrates})$ and $\textit{country}(\textit{X}, \textit{Y})$ result in the following variable substitutions after unification: $\{X/\textit{Greece}, Y/\textit{Socrates}\}$.
Prolog uses \emph{backward chaining} for proving assertions.
Given a goal atom $g$, this procedure first checks whether $g$ is explicitly stated in the KB -- in this case, it can be proven.
If it is not, the algorithm attempts to prove it by applying suitable rules, thereby generating subgoals that are proved next.
To find applicable rules, it attempts to unify $g$ with the heads of all available rules.
If this unification succeeds, the resulting variable substitutions are applied to the atoms in the rule body: each of those atoms becomes a subgoal, and each subgoal is recursively proven using the same strategy.
For instance, the application of the rule $\textit{country}(X, Y) \Leftarrow \textit{born\_in}(Y, X)$ to the goal $\textit{country}(\textit{Greece}, \textit{Socrates})$ would yield the subgoal $\textit{born\_in}(\textit{Socrates}, \textit{Greece})$.
Then the process is repeated for all subgoals until no subgoal is left to be proven.
The result of this procedure is a set of rule applications and variable substitutions referred to as \emph{proof}.
Note that the number of possible proofs grows exponentially with its depth, as every rule might be used in the proof of each subgoal.
Pseudo code for weak unification can be found in Appendix~\ref{app:algorithms} -- we refer the reader to \cite{russell2010artificial} for an in-depth treatment of the unification procedure.
\section{NLProlog} \label{sec:nlprolog}
Applying a logic reasoner to QA requires transforming the natural language paragraphs to logical representations, which is a brittle and error-prone process.

\begin{figure*}[t]
\centering
\includegraphics[width=0.6\linewidth]{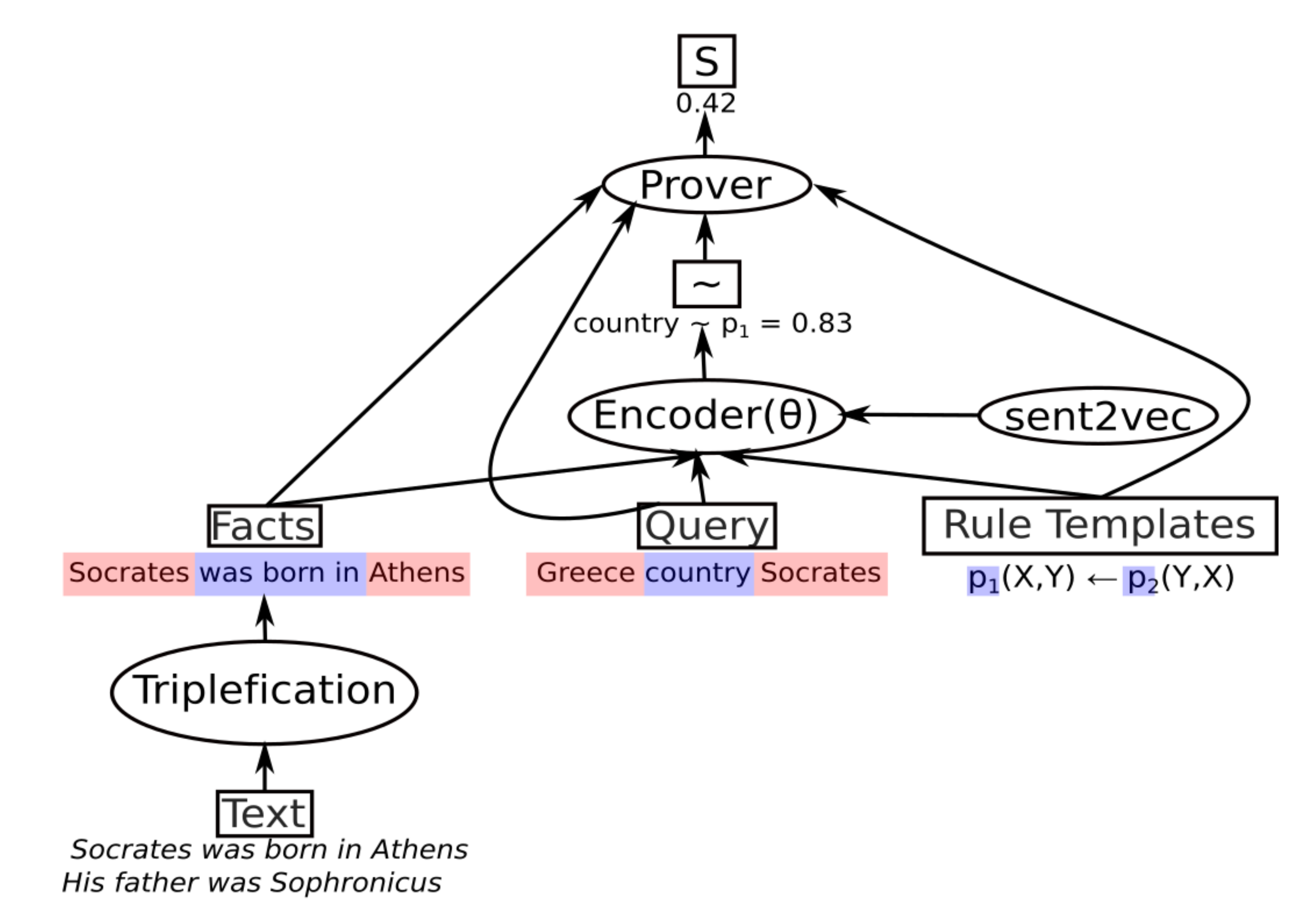}
\caption{Overview of \textsc{NLProlog} -- all components are depicted as ellipses, while inputs and outputs are drawn as squares. Phrases with red background are entities and blue ones are predicates.} \label{fig:overview}
\end{figure*}

Our aim is reasoning with natural language representations in the form of triples, where entities and relations may appear under different surface forms.
For instance, the textual mentions \emph{is located in} and \emph{lies in} express the same concept.
We propose replacing the exact matching between symbols in the Prolog unification operator with a \emph{weak unification} operator~\citep{sessa2002approximate}, which allows to unify two different symbols $s_{1}, s_{2}$, by comparing their representations using a differentiable similarity function $s_{1} \sim_\theta s_{2} \in [0,1]$ with parameters $\theta$.
With the weak unification operator, the comparison between two logical atoms results in an \emph{unification score} resulting from the aggregation of each similarity score.
Inspired by fuzzy logic t-norms~\citep{Gupta:1991:TTN:107687.107690}, aggregation operators are \eg{} the minimum or the product of all scores.
The result of backward-chaining with weak unification is a set of proofs, each associated with a proof score measuring the truth degree of the goal with respect to a given proof.
Similarly to backward chaining, where only successful proofs are considered, in \textsc{NLProlog} the final proof success score is obtained by taking the maximum over the success scores of all found proofs.
\textsc{NLProlog} combines inference based on the weak unification operator and distributed representations, to allow reasoning over sub-symbolic representations -- such as embeddings -- obtained from natural language statements.
Each natural language statement is first translated into a \emph{triple}, where the first and third element denote the entities involved in the sentence, and the second element denotes the \emph{textual surface pattern} connecting the entities.
All elements in each triple -- both the entities and the textual surface pattern -- are then embedded into a vector space. These vector representations are used by the similarity function $\sim_\theta$ for computing similarities between two entities or two textual surface patterns and, in turn, by the backward chaining algorithm with the weak unification operator for deriving a proof score for a given assertion.
Note that the resulting proof score is fully end-to-end differentiable with respect to the model parameters $\theta$: we can train \textsc{NLProlog} using gradient-based optimisation by back-propagating the prediction error to $\theta$.
\cref{fig:overview} shows an outline of the model, its components and their interactions.
\subsection{Triple Extraction} \label{ssec:triple_extraction}
To transform the support documents to natural language triples, we first detect entities by performing entity recognition with \textsc{spaCy}~\cite{spacy2}.
From these, we generate triples by extracting all entity pairs that co-occur in the same sentence and use the sentence as the predicate blinding the entities.
For instance, the sentence ``Socrates was born in Athens and his father was Sophronicus'' is converted in the following triples:
\begin{inparaenum}[\itshape i)\upshape]
\item (\emph{Socrates}, \emph{ENT1 was born in ENT2 and his father was Sophronicus}, \emph{Athens}),
\item (\textit{Socrates}, \textit{ENT1 was born in Athens and his father was ENT2}, \textit{Sophronicus}), and
\item (\textit{Athens}, \textit{Socrates was born in ENT1 and his father was ENT2}, \textit{Sophronicus}).
\end{inparaenum}
We also experimented with various Open Information Extraction frameworks~\citep{DBLP:conf/coling/NiklausCFH18}: in our experiments, such methods had very low recall, which led to significantly lower accuracy values.
\subsection{Similarity Computation}
Embedding representations of the symbols in a triple are computed using an encoder $\emb_\theta : \mathcal{F} \cup \mathcal{P} \mapsto \mathbb{R}^{d}$ parameterized by $\theta$ -- where $\mathcal{F}, \mathcal{P}$ denote the sets of entity and predicate symbols, and $d$ denotes the embedding size.
The resulting embeddings are used to induce the similarity function $\sim_{\theta} : (\mathcal{F} \cup \mathcal{P})^2 \mapsto [0, 1]$, given by their cosine similarity scaled to $\left[ 0, 1\right]$:
\begin{equation} \label{eq:sim}
s_1 \sim_{\theta{}} s_2 = \frac{1}{2} \left( 1 + \frac{\emb_{\theta}(s_1)^\top \emb_{\theta}(s_2)}{||\emb_{\theta}(s_1)|| \cdot ||\emb_{\theta}(s_2)||} \right)
\end{equation}
In our experiments, for using textual surface patterns, we use a sentence encoder composed of a static pre-trained component -- namely, \textsc{Sent2vec}~\citep{pagliardini2017unsupervised} -- and a Multi-Layer Perceptron (MLP) with one hidden layer and Rectified Linear Unit (ReLU) activations~\citep{DBLP:conf/iccv/JarrettKRL09}.
For encoding predicate symbols and entities, we use a randomly initialised embedding matrix.
During training, both the MLP and the embedding matrix are learned via backpropagation, while the sentence encoder is kept fixed.
Additionally, we introduce a third lookup table and MLP for the predicate symbols of rules and goals.
The main reason of this choice is that semantics of goal and rule predicates may differ from the semantics of fact predicates, even if they share the same surface form.
For instance, the query \triple{X}{parent}{Y} can be interpreted either as \triple{X}{is the parent of}{Y} or as \triple{X}{has parent}{Y}, which are semantically dissimilar.

\subsection{Training the Encoders} \label{ssec:training}
We train the encoder parameters $\theta$ on a downstream task via gradient-based optimization.
Specifically, we train \textsc{NLProlog} with backpropagation using a learning from entailment setting~\citep{muggleton1994inductive}, in which the model is trained to decide whether a Prolog program $\mathcal{R}$ entails the truth of a candidate triple $c \in C$, where $C$ is the set of candidate triples.
The objective is a model that assigns high probabilities $p(c|\mathcal{R};\theta)$ to true candidate triples, and low probabilities to false triples.
During training, we minimize the following loss:
\begin{equation} \label{eq:loss}
\begin{aligned}
L(\theta) = & - \log p(a|\mathcal{R};\theta) \\
& - \log\left( 1 - \max_{c \in C \setminus \{a\}} p(c|\mathcal{R};\theta) \right),
\end{aligned}
\end{equation}
where $a \in C$ is the correct answer.
For simplicity, we assume that there is only one correct answer per example, but an adaptation to multiple correct answers would be straight-forward, \eg{} by taking the minimum of all answer scores.
To estimate $p(c|\mathcal{R};\theta)$, we enumerate all proofs for the triple $c$ up to a given depth $D$, where $D$ is a user-defined hyperparameter.
This search yields a number of proofs, each with a success score $S_i$.
We set $p(c|\mathcal{R};\theta)$ to be the maximum of such proof scores:
\begin{equation*}
p(c|\mathcal{R};\theta) = S_{\max} = \max_{i} S_{i} \in [0, 1].
\end{equation*}
Note that the final proof score $p(c|\mathcal{R};\theta)$ only depends on the proof with maximum success score $S_{\max}$.
Thus, we propose to first conduct the proof search by using a prover utilizing the similarity function induced by the current parameters $\sim_{\theta_t}$, which allows us to compute the maximum proof score $S_{\max}$.
The score for each proof is given by the aggregation -- either using the \emph{minimum} or the \emph{product} functions -- of the weak unification scores, which in turn are computed via the differentiable similarity function $\sim_{\theta}$.
It follows that $p(c|\mathcal{R};\theta)$ is end-to-end differentiable, and can be used for updating the model parameters $\theta$ via Stochastic Gradient Descent.
\subsection{Runtime Complexity of Proof Search}\label{ssec:runtime}
The worst case complexity vanilla logic programming is exponential in the depth of the proof~\citep{russell2010artificial}.
However, in our case, this is a particular problem because 
weak unification requires the prover to attempt unification between all entity and predicate symbols.
To keep things tractable, \textsc{NLProlog} only attempts to unify symbols with a similarity greater than some user-defined threshold $\lambda$.
Furthermore, in the search step for one statement $q$, for the rest of the search, $\lambda$ is set to $\max(\lambda, S)$  whenever a proof for $q$ with success score $S$ is found.
Due to the monotonicity of the employed aggregation functions, this allows to prune the search tree without losing the guarantee to find the proof yielding the maximum success score $S_{\max}$, provided that $S_{\max} \geq \lambda$. 
We found this optimization to be crucial to make the proof search scale on the considered data sets.
\subsection{Rule Learning}
In \textsc{NLProlog}, the reasoning process depends on rules that describe the relations between predicates.
While it is possible to write down rules involving natural language patterns, this approach does not scale.
Thus, we follow \citet{rocktaschel2017end} and use rule templates to perform Inductive Logic Programming (ILP)~\citep{muggleton1991inductive}, which allows \textsc{NLProlog} to learn rules from training data.
In this setting, a user has to define a set of rules with a given structure as input.
Then, \textsc{NLProlog} can learn the rule predicate embeddings from data by minimizing the loss function in \cref{eq:loss} using gradient-based optimization methods.
For instance, to induce a rule that can model transitivity, we can use a rule template of the form $p_1(X, Z) \Leftarrow p_2(X, Y) \wedge p_3(Y, Z)$, and \textsc{NLProlog} will instantiate multiple rules with randomly initialized embeddings for $p_1$, $p_2$, and $p_3$, and fine-tune them on a downstream task.
The exact number and structure of the rule templates is treated as a hyperparameter.
Unless explicitly stated otherwise, all experiments were performed with the same set of rule templates containing two rules for each of the forms $q(X, Y) \Leftarrow p_{2}(X, Y)$, $p_{1}(X, Y) \Leftarrow p_{2}(Y, X)$ and $p_{1}(X, Z) \Leftarrow p_{2}(X, Y) \wedge p_{3}(Y, Z)$, where $q$ is the query predicate.
The number and structure of these rule templates can be easily modified, allowing the user to incorporate additional domain-specific background knowledge, such as $\textit{born\_in}(X, Z) \Leftarrow \textit{born\_in}(X, Y) \wedge \textit{located\_in}(Y, Z)$

\section{Evaluation}

We evaluate our method on two QA datasets, namely \textsc{MedHop}, and several subsets of \textsc{WikiHop}~\citep{welbl2017constructing}.
These data sets are constructed in such a way that it is often necessary to combine information from multiple documents to derive the correct answer.
In both data sets, each data point consists of a query $p(e, X)$, where $e$ is an entity, $X$ is a variable -- representing the entity that needs to be predicted, $C$ is a list of candidates entities, $a \in C$ is an answer entity and $p$ is the query predicate.
Furthermore, every query is accompanied by a set of support documents which can be used to decide which of the candidate entities is the correct answer.
\subsection{MedHop} \label{ssec:MedHop}
\textsc{MedHop} is a challenging multi-hop QA data set, and contains only a single query predicate.
The goal in \textsc{MedHop} is to predict whether two drugs interact with each other, by considering the interactions between proteins that are mentioned in the support documents.
Entities in the support documents are mapped to data base identifiers.
To compute better entity representations, we reverse this mapping and replace all mentions with the drug and proteins names gathered from \textsc{DrugBank}~\citep{wishart2006drugbank} and \textsc{UniProt}~\citep{apweiler2004uniprot}.
\subsection{Subsets of WikiHop} \label{ssec:WikiHop}
To further validate the effectiveness of our method, we evaluate on different subsets of \textsc{WikiHop}~\citep{welbl2017constructing}, each containing a single query predicate.
We consider the predicates \textit{publisher}, \textit{developer}, \textit{country}, and \textit{record\_label}, because their semantics ensure that the annotated answer is unique and they contain a relatively large amount of questions that are annotated as requiring multi-hop reasoning.
For the predicate \textit{publisher}, this yields 509 training and 54 validation questions, for \textit{developer} 267 and 29, for \textit{country} 742 and 194, and for \textit{record\_label} 2305 and 283.
As the test set of \textsc{WikiHop} is not publicly available, we report scores for the validation set.
\subsection{Baselines} \label{ssec:baselines}
Following \citet{welbl2017constructing}, we use two neural QA models, namely \textsc{BiDAF}~\citep{seo2016bidirectional} and \textsc{FastQA}~\citep{weissenborn2017fastqa}, as baselines for the considered \textsc{WikiHop} predicates.
We use the implementation provided by the \textsc{Jack}~\footnote{\url{https://github.com/uclmr/jack}} QA framework~\citep{weissenborn2018jack} with the same hyperparameters as used by \citet{welbl2017constructing}, and train a separate model for each predicate.\footnote{We also experimented with the AllenNLP implementation of \textsc{BiDAF}, available at \url{https://github.com/allenai/allennlp/blob/master/allennlp/models/reading_comprehension/bidaf.py}, obtaining comparable results.}
To ensure that the performance of the baseline is not adversely affected by the relatively small number of training examples, we also evaluate the \textsc{BiDAF} model trained on the whole \textsc{WikiHop} corpus.
In order to compensate for the fact that both models are extractive QA models which cannot make use of the candidate entities, we additionally evaluate modified versions which transform both the predicted answer and all candidates to vectors using the \textit{wiki-unigrams} model of \textsc{Sent2vec}~\citep{pagliardini2017unsupervised}.
Consequently, we return the candidate entity which has the highest cosine similarity to the predicted entity.
We use the normalized version of \textsc{MedHop} for training and evaluating the baselines, since we observed that denormalizing it (as for \textsc{NLProlog}) severely harmed performance.
Furthermore on \textsc{MedHop}, we equip the models with word embeddings that were pretrained on a large biomedical corpus~\citep{Pyysalo2013DistributionalSR}.

\subsection{Hyperparameter Configuration}
\label{ssec:hyperparams}
On \textsc{MedHop} we optimize the embeddings of predicate symbols of rules and query triples, as well as of entities.
\textsc{WikiHop} has a large number of unique entity symbols and thus, learning their embeddings is prohibitive.
Thus, we only train the predicate symbols of rules and query triples on this data set.
For \textsc{MedHop} we use bigram \textsc{Sent2vec} embeddings trained on a large biomedical corpus~\footnote{\url{https://github.com/ncbi-nlp/BioSentVec}}, and for \textsc{WikiHop} the \textit{wiki-unigrams} model\footnote{\url{https://drive.google.com/open?id=0B6VhzidiLvjSa19uYWlLUEkzX3c}} of \textsc{Sent2vec}.
All experiments were performed with the same set of rule templates containing two rules for each of the forms $p(X, Y) \Leftarrow q(X, Y)$, $p(X, Y) \Leftarrow q(Y, X)$ and $p(X, Z) \Leftarrow q(X, Y) \wedge r(Y, Z)$ and set the similarity threshold $\lambda$ to $0.5$ and maximum proof depth to $3$.
We use Adam~\citep{kingma2014adam} with default parameters.

\subsection{Results}
The results for the development portions of \textsc{WikiHop} and \textsc{MedHop} are shown in Table~\ref{tab:results_countryhop}.
For all predicates but \textit{developer}, \textsc{NLProlog} strongly outperforms all tested neural QA models, while achieving the same accuracy as the best performing QA model on \textit{developer}.
We evaluated \textsc{NLProlog} on the hidden test set of MedHop and obtained an accuracy of 29.3\%, which is 6.1 pp better than FastQA and 18.5 pp worse than BiDAF.\footnote{Note, that these numbers are taken from \citet{welbl2017constructing} and were obtained with different implementations of \textsc{BiDAF} and \textsc{FastQA}}.
As the test set is hidden, we cannot diagnose the exact reason for the inconsistency with the results on the development set, but observe that FastQA suffers from a similar drop in performance.
\begin{table*}[ht]
\centering
\begin{tabular}{l c c c c c }
    \toprule
    {\bf Model} & {\bf MedHop} & \textit{publisher} & \textit{developer} & \textit{country} & \textit{recordlabel} \\
    \midrule
    BiDAF & 42.98 & 66.67 & 65.52 & 53.09 & 68.90 \\
    \quad + Sent2Vec & --- & 75.93 & \textbf{68.97} & 61.86 & 75.62 \\
    \quad + Sent2Vec + wikihop & --- & 74.07 & 62.07 & 66.49 & 78.09 \\
    FastQA & 52.63 & 62.96 & 62.07 & 57.21 & 70.32 \\
    \quad + Sent2Vec & --- & 75.93 & 58.62 & 64.95 & 78.09 \\
    \midrule 
    NLProlog & \textbf{65.78} & \textbf{83.33} & \textbf{68.97} & \textbf{77.84} &  \textbf{79.51}\\
    \quad - rules & 64.33 & \textbf{83.33} & \textbf{68.97} & 74.23 & 74.91\\
    \quad - entity MLP & 37.13 & 68.52 & 41.38 & 72.16 & 64.66\\
    \bottomrule

\end{tabular}
\caption{Accuracy scores in percent for different predicates on the development set of the respective predicates. +/- denote independent modifications to the base algorithm.
}
\label{tab:results_countryhop}
\end{table*}
\subsection{Importance of Rules}
Exemplary proofs generated by \textsc{NLProlog} for the predicates \textit{record\_label} and \textit{country} can be found in \cref{fig:proof_trees}.

To study the impact of the rule-based reasoning on the predictive performance, we perform an ablation experiment in which we train \textsc{NLProlog} without any rule templates.
The results can be found in the bottom half of Table~\ref{tab:results_countryhop}.
On three of the five evaluated data sets, performance decreases markedly when no rules can be used and does not change on the remaining two data sets.
This indicates that reasoning with logic rules is beneficial in some cases and does not hurt performance in the remaining ones.

\subsection{Impact of Entity Embeddings}
\label{ssec:entity_mlp}
\begin{figure}[t]
    \centering
    \includegraphics[width=1\linewidth]{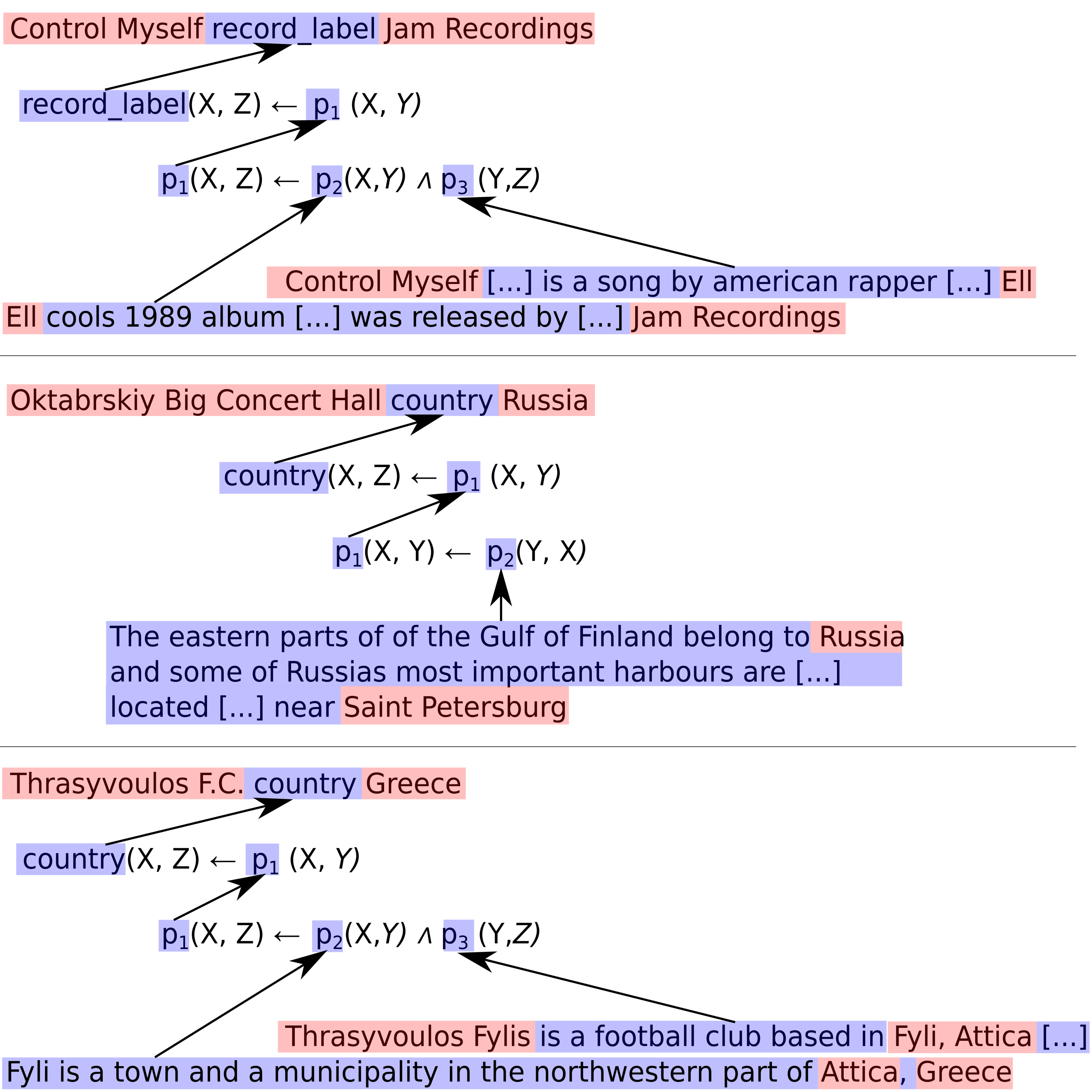}
    \caption{Example proof trees generated by \textsc{NLProlog}, showing a combination of multiple rules. Entities are shown in red and predicates in blue. Note, that entities do not need to match exactly.
    The first and third proofs were obtained without the entity MLP (as described in Section~\ref{ssec:entity_mlp}), while the second one was obtained in the full configuration of \textsc{NLProlog}.}
    \label{fig:proof_trees}
\end{figure}

In a qualitative analysis, we observed that in many cases multi-hop reasoning was performed via aligning entities and not by applying a multi-hop rule. 
For instance, the proof of the statement \textit{country(Oktabrskiy Big Concert Hall, Russia)} visualized in Figure~\ref{fig:proof_trees}, is performed by making the embeddings of the entities \textit{Oktabrskiy Big Concert Hall} and \textit{Saint Petersburg} sufficiently similar.
To gauge the extent of this effect, we evaluate an ablation in which we remove the MLP on top of the entity embeddings.
The results, which can be found in Table~\ref{tab:results_countryhop}, show that fine-tuning entity embeddings plays an integral role, as the performance degrades drastically.  
Interestingly, the observed performance degradation is much worse than when training without rules, suggesting that much of the reasoning is actually performed by finding a suitable transformation of the entity embeddings.

\subsection{Error Analysis}\label{ssec:error_analysis}
We performed an error analysis for each of the \textsc{WikiHop} predicates.
To this end, we examined all instances in which one of the neural QA models (with \textsc{Sent2Vec}) produced a correct prediction and \textsc{NLProlog} did not, and labeled them with pre-defined error categories.
Of the 55 instances, $49\%$ of the errors were due to \textsc{NLProlog} unifying the wrong entities, mainly because of an over-reliance on heuristics, such as predicting a record label if it is from the same country as the artist.
In $25\%$ of the cases, \textsc{NLProlog} produced a correct prediction, but another candidate was defined as the answer. In $22\%$ the prediction was due to an error in predicate unification, \ie{} \textsc{NLProlog} identified the correct entities, but the sentence did not express the target relation.
Furthermore, we performed an evaluation on all problems of the studied \textsc{WikiHop} predicates that were unanimously labeled as containing the correct answer in the support texts by ~\citet{welbl2017constructing}.
On this subset, the micro-averaged accuracy of \textsc{NLProlog} shows an absolute increase of $3.08$ pp, while the accuracy of \textsc{BiDAF} (\textsc{FastQA}) augmented with \textsc{Sent2Vec} decreases by $3.26$ ($3.63$) pp.
We conjecture that this might be due to \textsc{NLProlog}'s reliance on explicit reasoning, which could make it less susceptible to spurious correlations between the query and supporting text.

\section{Discussion and Future Work}
We proposed \textsc{NLProlog}, a system that is able to perform rule-based reasoning on natural language, and can learn domain-specific rules from data.
To this end, we proposed to combine a symbolic prover with pretrained sentence embeddings, and to train the resulting system using backpropagation.
We evaluated \textsc{NLProlog} on two different QA tasks, showing that it can learn domain-specific rules and produce predictions which outperform those of the two strong baselines \textsc{BiDAF} and \textsc{FastQA} in most cases.
While we focused on a subset of First Order Logic in this work, the expressiveness of \textsc{NLProlog} could be extended by incorporating a different symbolic prover.
For instance, a prover for temporal logic~\citep{Orgun1994-jt} would allow to model temporal dynamics in natural language.
We are also interested in incorporating future improvements of symbolic provers, triple extraction systems and pretrained sentence representations to further enhance the performance of \textsc{NLProlog}.
Additionally, it would be interesting to study the behavior of \textsc{NLProlog} in the presence of multiple \textsc{WikiHop} query predicates.

\section*{Acknowledgments}
Leon Weber and Jannes Münchmeyer acknowledge the support of the Helmholtz Einstein International Berlin Research School in Data Science (HEIBRiDS). We would like to thank the anonymous reviewers for the constructive feedback. We gratefully acknowledge the support of NVIDIA Corporation with the donation of a Titan X Pascal GPU used for this research.

\bibliography{bibliography,refs_clean_nourl}
\bibliographystyle{acl_natbib}

\newpage

\begin{appendices}
 
\section{Algorithms}
\label{app:algorithms}
\begin{algorithm}[h!]
    \SetAlgoLined\DontPrintSemicolon
    \SetKwProg{unify}{fun $\textit{unify}(x, y, \theta, S)$}{}{}
    \SetKwProg{unifyvar}{fun $\textit{unify\_var}(v, o, \theta, S)$}{}{}

    \unify{}{
        \KwIn{
            \\
            $x$: function $f(\ldots)$ \textpipe{} atom $p(\ldots)$ \textpipe{} \textit{variable} \textpipe{}  list $x_1 :: x_2 :: \ldots :: x_n$\\
            $y$: function $f'(\ldots)$ \textpipe{}  atom $p'(\ldots)$ \textpipe{} \textit{variable} \textpipe{}  list $y_1 :: y_2 :: \ldots :: y_m$\\
            $\theta$: current substitutions, default = $\{\}$\\
            $S$: current success score, default = $1.0$\\
            }
        \KwOut{(Unifying substitution $\theta'$ or \textit{failure}, Updated success score $S'$)}
        \lIf{$\theta = \textit{failure}$}{\Return (\textit{failure}, $0$)}
        \lElseIf{$S < \lambda$}{\Return (\textit{failure}, $0$)}
        \lElseIf{$x = y$}{\Return ($\theta$, $S$)}

        \lElseIf{$x$ is \textit{Var} }{\Return $\textit{unify\_var}(x, y, \theta, S)$}
        \lElseIf{$y$ is \textit{Var} }{\Return $\textit{unify\_var}(y, x, \theta, S)$}
        \ElseIf{$x$ is $f(x_1, \ldots, x_n)$, $y$ is $f'(y_1, \ldots, y_n)$, and $f \sim f' \geq \lambda$}{
             $S' := S \wedge f \sim f'$\\
             \Return $\textit{unify}(x_1 :: \ldots :: x_n, y_1 :: \ldots :: y_n, \theta, S')$
         }
        \ElseIf{$x$ is $p(x_1, \ldots, x_n)$, $y$ is $p'(y_1, \ldots, y_n)$, and $p \sim p' \geq \lambda$}{
             $S' := S \wedge f \sim f'$\\
             \Return $\textit{unify}(x_1 :: \ldots :: x_n, y_1 :: \ldots :: y_n, \theta, S')$
         }
        \ElseIf{$x$ is $x_1 :: \ldots :: x_n$ and y is $y_1 :: \ldots :: y_n$}{
             $(\theta', S') := \textit{unify}(x_1, y_1, \theta, S)$\\
             \Return $\textit{unify}(x_2 :: \ldots :: x_n, y_2 :: \ldots :: y_n, \theta', S')$
         }
         \lElseIf{$x$ is empty list and y is empty list}{\Return $(\theta, S)$}
        \lElse{\Return (\textit{failure}, $0$)}
 }

 \unifyvar{}{
     \lIf{$\{\textit{v}/\textit{val}\} \in \theta$}{\Return $\textit{unify}(\textit{val}, o, \theta, S)$}
     \lElseIf{$\{o/\textit{val}\} \in \theta$}{\Return $\textit{unify}(\textit{var}, \textit{val}, \theta, S)$}
     \lElse{\Return $(\{v/o\} + \theta, S)$}
 }
 \caption{The weak unification algorithm in \textsc{NLProlog} without occurs check}
 \label{alg:weak_unification}
\end{algorithm}

\end{appendices}

\end{document}